\documentclass[sigconf]{acmart}
\usepackage{listings}

\usepackage{color}
\usepackage{amsmath}
\usepackage{graphicx}

\DeclareFixedFont{\ttb}{T1}{txtt}{bx}{n}{7} 
\DeclareFixedFont{\ttm}{T1}{txtt}{m}{n}{7}  

\definecolor{deepblue}{rgb}{0,0,0.5}
\definecolor{deepred}{rgb}{0.6,0,0}
\definecolor{deepgreen}{rgb}{0,0.5,0}
\newcommand\pythonstyle{\lstset{
language=Python,
basicstyle=\ttm,
morekeywords={self},              
keywordstyle=\ttb\color{deepblue},
emph={MyClass,__init__},          
emphstyle=\ttb\color{deepred},    
stringstyle=\color{deepgreen},
frame=tb,                         
showstringspaces=false
}}

\lstnewenvironment{python}[1][]
{
\pythonstyle
\lstset{#1}
}
{}

\newcommand\pythoninline[1]{{\pythonstyle\lstinline!#1!}}

\AtBeginDocument{%
  \providecommand\BibTeX{{%
    \normalfont B\kern-0.5em{\scshape i\kern-0.25em b}\kern-0.8em\TeX}}}

\copyrightyear{2021}
\acmYear{2021}
\setcopyright{iw3c2w3}
\acmConference[WWW '21 Companion]{Companion Proceedings of the Web Conference 2021}{April 19--23, 2021}{Ljubljana, Slovenia}
\acmBooktitle{Companion Proceedings of the Web Conference 2021 (WWW '21 Companion), April 19--23, 2021, Ljubljana, Slovenia}
\acmPrice{}
\acmDOI{10.1145/3442442.3458602}
\acmISBN{978-1-4503-8313-4/21/04}

\begin{document}

\title{TagRuler: Interactive Tool for Span-Level Data Programming by Demonstration}

\author{Dongjin Choi}
\authornote{Work done during internship at Megagon Labs.}
\affiliation{%
  \country{Georgia Institute of Techonology, USA}
}
\email{jin.choi@gatech.edu}

\author{Sara Evensen}
\affiliation{%
  \country{Megagon Labs, USA}
}
\email{sara@megagon.ai}

\author{Çağatay Demiralp}
\authornote{Work done when at Megagon Labs.}
\affiliation{%
  \country{Sigma Computing, USA}
}
\email{cagatay@sigmacomputing.com}

\author{Estevam Hruschka}
\affiliation{%
  \country{Megagon Labs, USA}
}
\email{estevam@megagon.ai}

\renewcommand{\shortauthors}{Choi et al.}

\begin{abstract}
Despite rapid developments in the field of machine learning research, collecting high quality labels for supervised learning remains a bottleneck for many applications.  This difficulty is exacerbated by the fact that state-of-the art models for NLP tasks are becoming deeper and more complex, often increasing the amount of training data required even for fine-tuning.
Weak supervision methods, including data programming, address this problem and reduce the cost of label collection by using noisy label sources for supervision.  However until recently, data programming was only accessible to users who knew how to program.
In order to bridge this gap, the Data Programming by Demonstration framework was proposed to facilitate the automatic creation of labeling functions based on a few examples labeled by a domain expert.
This framework has proven successful for generating high accuracy labeling models for document classification. 
In this work, we extend the DPBD framework to span-level annotation tasks, arguably one of the most time consuming NLP labeling tasks. We built a novel tool, TagRuler, that makes it easy for annotators to build span-level labeling functions without programming and encourages them to explore trade-offs between different labeling models and active learning strategies. We empirically demonstrated that an annotator could achieve a higher F1 score using the proposed tool compared to manual labeling for different span-level annotation tasks.
\end{abstract}

%
%




\maketitle
\definecolor{orange}{RGB}{255,119,0}
\definecolor{red}{RGB}{220,0,0}
\definecolor{agreen}{RGB}{74, 198, 148}
\definecolor{purple}{RGB}{158, 62, 177}
\definecolor{darkpurple}{RGB}{170, 70, 210}
\definecolor{aqua}{RGB}{87, 180, 181}
\definecolor{lightblue}{RGB}{72, 123, 232}
\definecolor{hotpink}{RGB}{255, 83, 115}
\definecolor{teal}{RGB}{90, 200, 250}
\definecolor{linkColor}{RGB}{6,125,233}
\definecolor{tomato}{rgb}{1,0.2,0}
\definecolor{grey}{rgb}{0.4,0.4,0.4}

\newcommand{\todo}[1]{\textcolor{hotpink}{[TODO : #1]}}
\newcommand{\estevam}[1]{\small[\textcolor{grey}{Estevam:}\textcolor{red}{#1}]}
\newcommand{\jin}[1]{\small[\textcolor{grey}{Jin:}\textcolor{aqua}{#1}]}
\newcommand{\sara}[1]{\small[\textcolor{grey}{Sara:}\textcolor{orange}{#1}]}
\newcommand{\cagatay}[1]{\small[\textcolor{grey}{\c{C}a\u{g}atay:}\textcolor{tomato}{#1}]}

\newcommand{\mkclean}{
  \renewcommand{\jin}[1]{}
  \renewcommand{\sara}[1]{}
  \renewcommand{\estevam}[1]{}
  \renewcommand{\todo}[1]{}
}

\makeatletter
\newcommand{\oi}{\mathbin{\mathpalette\make@circled i}}
\newcommand{\make@circled}[2]{%
  \ooalign{$\m@th#1\smallbigcirc{#1}$\cr\hidewidth$\m@th#1#2$\hidewidth\cr}%
}
\newcommand{\smallbigcirc}[1]{%
  \vcenter{\hbox{\scalebox{1.3}{$\m@th#1\bigcirc$}}}%
}
\makeatother

\vspace{-2mm}
\section{Introduction}
In recent machine learning research, supervised learning has remained a dominant approach in situations where accuracy is critical. This is despite the increasing complexity of models, which in turn requires more training data, even when taking advantage of pre-trained models (as they frequently still require large refined datasets for supervised fine-tuning).
In many cases, labels are manually annotated by humans and can require from tens to hundreds of thousands of examples~\cite{ratner2017snorkel}.
For example, one of the state-of-the-art open domain named entity recognition models~\cite{li2019dice}, used more than a million data samples.
Thus, the need for big amounts of labeled data and subsequent human efforts are still a big bottleneck in the recent NLP field.
\vspace{-1mm}
\paragraph{\textbf{Weak Supervision and Data Programmming}}
One common approach to reduce the need for labeled data is to rely on weak supervision methods~\cite{mintz2009distant,hoffmann2011knowledge}. Such methods suggest using noisy (thus, cheaper) label sources for supervision to reduce the labeling cost. Crowd-sourcing~\cite{karger2011iterative} and user-defined heuristics~\cite{gupta2014improved} are among frequently used label sources to assign noisy training labels to available unlabeled data. 
Another alternative, which is gaining a lot of attention in recent years is data programming (DP) \cite{ratner2017snorkel}.
In particular, DP focuses on the process of generating domain knowledge from domain experts in the form of labeling functions using programming language.
\begin{python}
def lf(x):
    if "http" in x:     return SPAM
    else:               return ABSTAIN
\end{python}
The above code snippet shows a simple example of a labeling function (written in Python) for spam classification task.
A main limitation of data programming is that domain experts for many difficult natural language tasks (e.g. medical document annotation, legal document annotation, etc.) may not have experience in programming, making it difficult for them to create such labeling functions.
\vspace{-4mm}
\paragraph{\textbf{Data Programmming by Demonstration (DPBD)}}
Aiming at bridging the gap between domain experts and labeling functions creation, Ruler~\cite{evensen2020ruler} proposed the concept of Data Programming by Demonstration.
DPBD is a human-in-the-loop model for synthesizing labeling functions through interactive user feedback, eliminating the need for domain experts to learn programming skills to leverage data programming.
In this paper, we introduce TagRuler, an open-source web application system that extends the DPBD framework to span-level annotation tasks.
TagRuler includes a set of new features and characteristics that are not explored before in Ruler. 
More specifically, in TagRuler we adapted the DPBD main components to fit the span-level annotation problem and added new models. TagRuler implements four different models and allows users to explore the trade-offs afforded by each.

We designed TagRuler to optimize usability by carefully selecting atomic conditions for span labeling rules.
Based on results previously described in the literature 
(e.g. BabbleLabble~\cite{hancock2018training}),  lexical, semantic, syntactic, and entity-type match rules were incorporated into TagRuler, in addition to the possibility of using the negation operator. 
We implement semantic rules via similarity based on contextual token embeddings (BERT~\cite{devlin2018bert} and ELMo \cite{peters2018deep}).
In addition to the active learning approach previously implemented in Ruler, TagRuler proposes a novel active learning component focusing on false positives, thus contributing to one of the main challenges in data programming: surfacing difficult (or borderline) labeled examples~\cite{suri2020leveraging}.
In this active learning approach, unlabeled examples that have higher potential to identify false positives will have higher probability to be sampled as next instance to be labeled. It aims to achieve maximum accuracy with a small number of interactions combined with uncertainty-based sampling.


\paragraph{\textbf{Contributions}}

This work presents the following contributions:
\begin{itemize}
    \item proposes a token labeling model which incorporates logical combinations of expressive atomic rules.
    \item implements state-of-the-art sequential label aggregation models.
    \item adds an active learning component based on false positives.
    \item demonstrates the effectiveness of TagRuler through evaluation on use-cases from two domains.
    \item makes the underlying code\footnote{Code available at \url{https://github.com/megagonlabs/tagruler}.}, example data, and video demo\footnote{A video demo can be found at \url{https://youtu.be/MRc2elPaZKs}.}
    publicly available in order to make span-level data programming for text more accessible to researchers and annotators. 
\end{itemize}

\begin{figure}[t]
    \centering
    \includegraphics[width=0.45\textwidth]{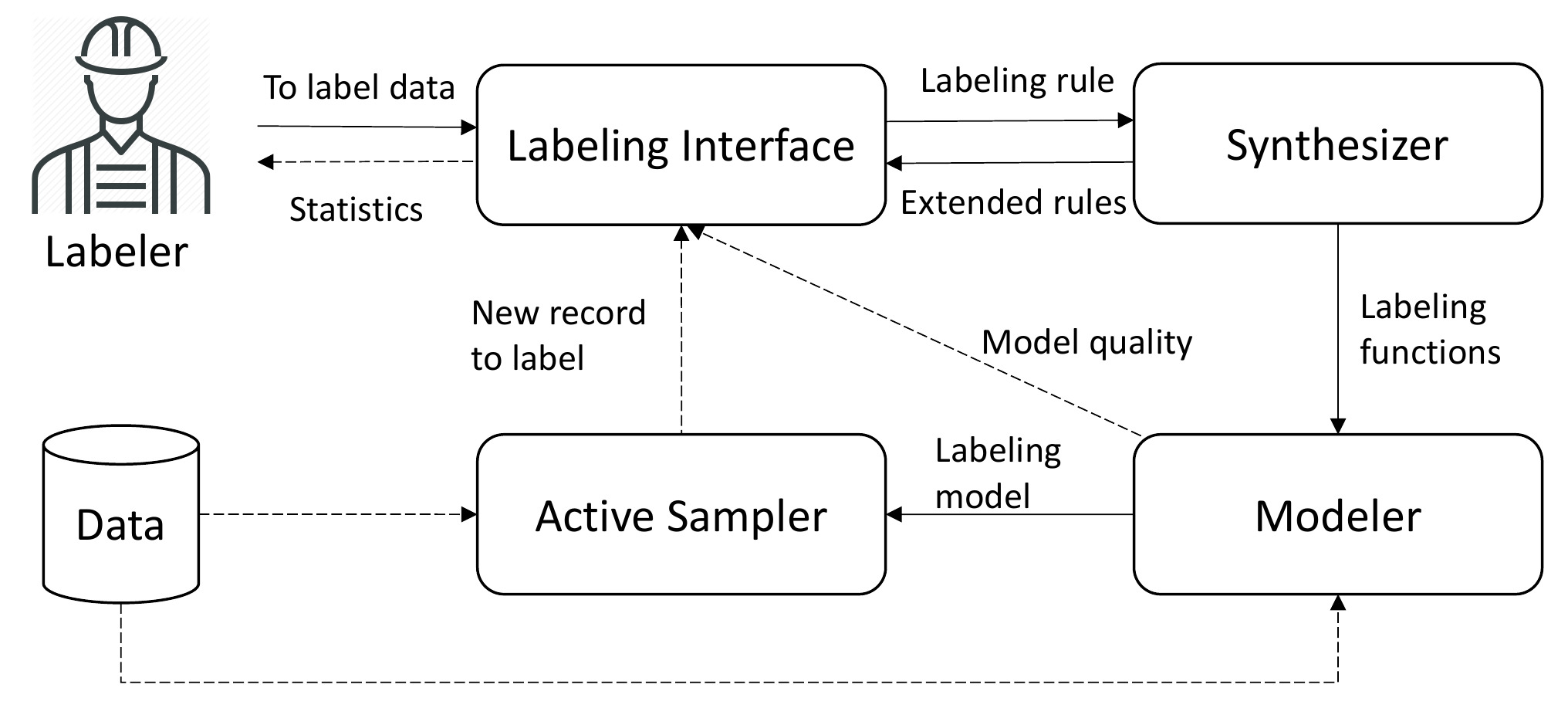}
    \vspace{-3mm}
    \caption{The Data Programming By Demonstration (DPBD) framework. Straight lines indicate the flow of knowledge, and dashed lines indicate the flow of data.}
    \vspace{-5mm}
    \label{fig:architecture}
\end{figure}
\begin{figure*}[t]
\centering
\includegraphics[width=0.93\textwidth]{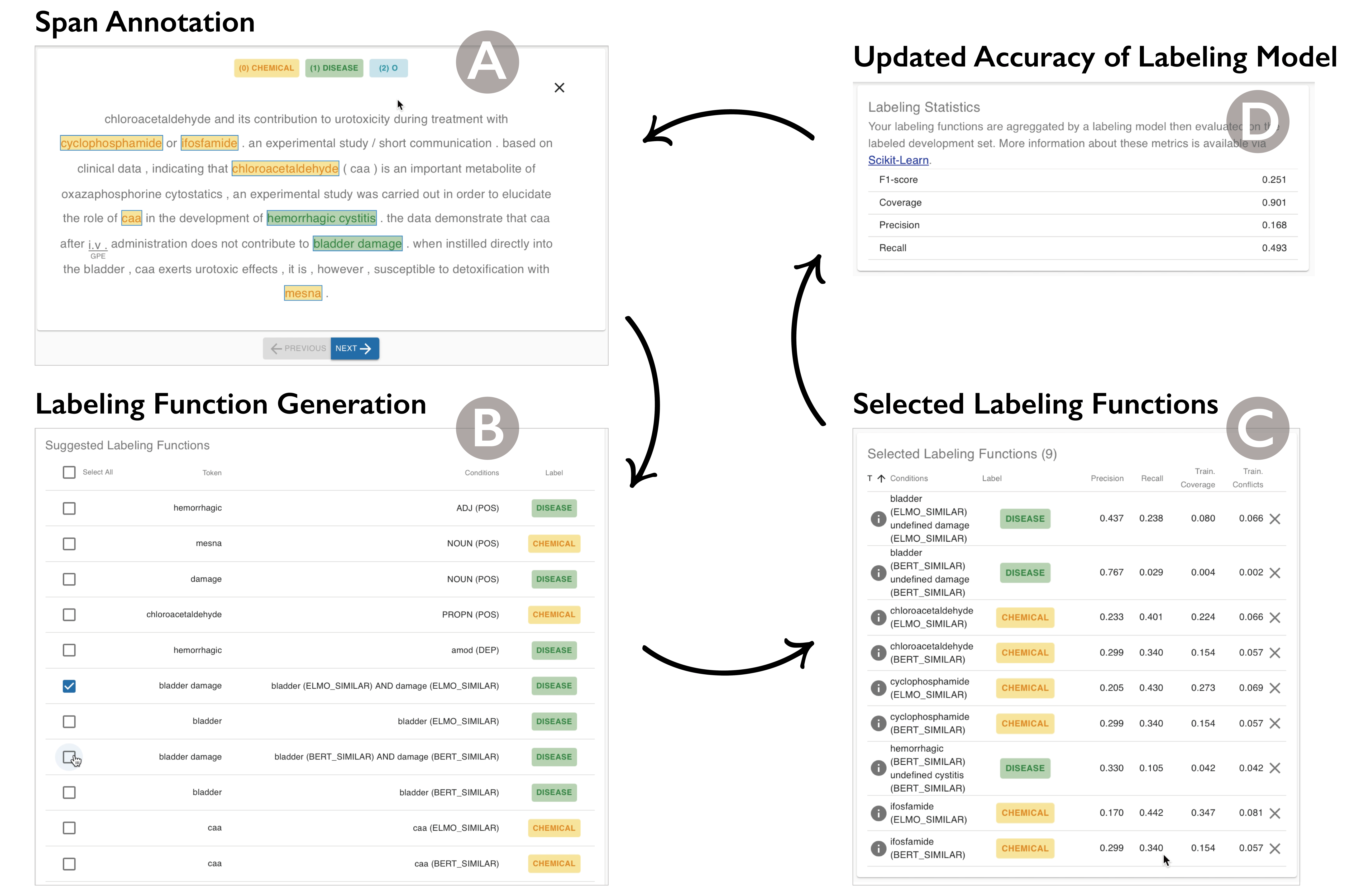}
\vspace{-1mm}
\caption{Overview of TagRuler. (A) Main text view lets users annotate text spans with relevant class names.
(B) Labeling function view shows a list of labeling functions suggested by TagRuler. The user is expected to select relevant labeling functions that align with their domain knowledge.
(C) Selected labeling function view shows statistics about the user-selected functions.
(D) Label model view shows statistics of the label model trained on currently selected labeling functions.}
\vspace{-1mm}
\label{fig:overview}
\end{figure*}
\vspace{-2mm}
\section{System Overview}
\autoref{fig:architecture} shows the architecture of DPBD framework that was proposed by ~\cite{evensen2020ruler}.
In this section, we describe how TagRuler extends each component of the DPBD framework for span-level annotation, or tagging task.
We also describe how each component is connected to the TagRuler's user interface shown in~\autoref{fig:overview}.
\vspace{-1mm}
\subsection{Labeling Interface}
TagRuler provides a web-based interactive system that facilitates users to encode their domain knowledge for span annotation into labeling rules.
To this end, TagRuler provides the main text view (\autoref{fig:overview} (A)) where user interacts with a text data sample.
The main text view shows a single text data and supports labeling operations.
The user annotates spans by highlighting them with the cursor.
Next, one assigns a label to the selected spans by choosing one from the options displayed.
The user also can indicate if the selected span is a positive or negative sample by clicking the (+) or (-) signs that pop up as a span is selected.
Once labels are assigned to selected spans, the \textit{synthesizer} compiles the selections and generates a set of relevant labeling functions.

\subsection{Synthesizer}
The synthesizer translates the annotations into atomic rules
and generates a candidate set of labeling functions (in programming language) using the rules.
TagRuler implements atomic rules which are specifically for span annotation tasks.
Following previous works on rule-based labeling models~\cite{hancock2018training, safranchik2020weakly}, we focus on capturing semantic, syntactic, and entity-type information of the spans.
The created labeling functions are shown in the labeling function view~(\autoref{fig:overview} (B)), and the user selects relevant labeling functions and delivers them to the model.
The selected labeling functions are shown in ~\autoref{fig:overview} (C) with corresponding statistics (accuracy and coverage) of each labeling function.
Lastly they are aggregated by the \textit{modeler} and used to produce accuracy feedback in an iterative process.

\subsubsection{Contextual similarity rules with neural embeddings}
For semantic similarity, we implement two dense vector representation (embeddings) rules, \verb|SIMILAR_BERT()| and \verb|SIMILAR_ELMO()| rules, which are based on BERT~\cite{devlin2018bert} and ELMo~\cite{peters2018deep}, respectively. Both return True if the embeddings of the two input tokens have higher cosine similarity then pre-defined threshold values.
We use the pretrained models since they are able to capture contextual semantics considering the structure of sentences implicitly.
During demonstration, we observed that \verb|SIMILAR_BERT()| and \verb|SIMILAR_ELMO()| work differently and are complementary.
As BERT has been the state-of-the-art contextual language model, it has more strength in capturing semantic and syntactic information implicitly at the same time.
ELMo outputs more stable results for out-of-vocabulary words since it is a character-based model.
This functionality is especially important when a user works on highly domain-specific data sets since they tend to include rare vocabularies that were under-trained in the BERT model. We observed this effect when testing TagRuler on a medical dataset.

\subsubsection{POS, dependency parse, and NER tag rules}
Syntactic information of word is also important to build expressive rules.
Two representative ways of extracting the syntactic information are Part-of-Speech (POS) tagging and dependency parsing, which capture shallow and deeper syntactic structure, respectively.
Specifically, we implement \verb|POS()| and \verb|DEP()| rules.
Both return True if two input tokens have same POS and DEP tags, respectively.
Similarly, TagRuler supports usage of \verb|NER()| rule which returns True if inputs have a same named entity recognition (NER) tag (in our case, 18 classes provided by spaCy~\cite{spacy2})
To detect POS, dependency, and NER tags, we used the spaCy Python library~\cite{spacy2} which have achieved near-human performance for each task.
\vspace{-2mm}
\subsection{Modeler}
The modeler creates weak supervision labels and associated probabilities by using the output of labeling functions selected by the user.
That is, given a set of labels $\mathcal{L}$, and a label matrix $M \in \{1, 2, \dots, |\mathcal{L}| \}^{n \times l}$ generated by $l$ labeling functions applied to $n$ tokens, the modeler fits an aggregation model and produce a matrix $P \in \mathbb{R}^{n \times |\mathcal{L}|}$ where $P_{i,j}$ denotes the probability $P(label(token_i)== j | token_i)$.
The resulting probabilistic labels are applied to a small labeled development set, and their accuracy is shown to the user in the label model view (\autoref{fig:overview}(D)) in real time.
TagRuler implements four different models: Snorkel~\cite{ratner2017snorkel}, HMM~\cite{safranchik2020weakly}, FlyingSquid~\cite{fu2020fast}, and Majority Voter to aggregate and generate probabilistic labels.
Snorkel, HMM, and FlyingSquid are probabilistic graphical models that capture dependency structure between tokens and labeling functions.
In particular, HMM and FlyingSquid support sequential dependence structure, while others do not.
It is important to use the appropriate method as modeler, since there is a tradeoff between the time it takes to fit the model (which hampers interactivity) and the accuracy.
At the beginning of a session, users can select the model most appropriate for their use-case.
\subsection{Active Sampler}
TagRuler is a system designed to learn from expert knowledge, and expert manual annotation is expensive, so it is important to obtain informative data with as few annotations as possible.
The active learning approach focuses on contributing to this main challenge in data programming: generating difficult (or borderline) examples~\cite{suri2020leveraging}. 
TagRuler samples the text to be displayed in \autoref{fig:overview} (A) after each annotation using an active learning technique that leverages the trained label model and a small labeled development set.

TagRuler's active learning approach is motivated by the fact that there is a small set of ground truth in the DPBD setting.
In this approach, unlabeled  examples  that  have  higher  potential  in  identifying and reducing false positives will have higher probability to be sampled as next instance to be labeled.
First, the active sampler selects the data record $x^*$ from the development set, which shows the greatest error in predicting the true label, i.e.\ $x^{*} = \arg\min_{x} p(label_{true}(x)|x)$.
Then, the sampler samples the most similar text from train set using cosine similarity between sentence embedding vectors.
For sentence embeddings, we use the Sentence-bert~\cite{reimers2019sentence} model. This model is based on the BERT model so that the embeddings can be comprehensive representation of syntax and semantics of texts. 
TagRuler implements an active sampler that alternately samples based on the proposed method and the uncertainty-based sampler used by Ruler~\cite{evensen2020ruler} so that it can advantage of both methods in turn, reducing variance as \textit{ensemble models} do.

\begin{figure}[h]
    \centering
    \vspace{-1mm}
    \includegraphics[width=0.46\textwidth]{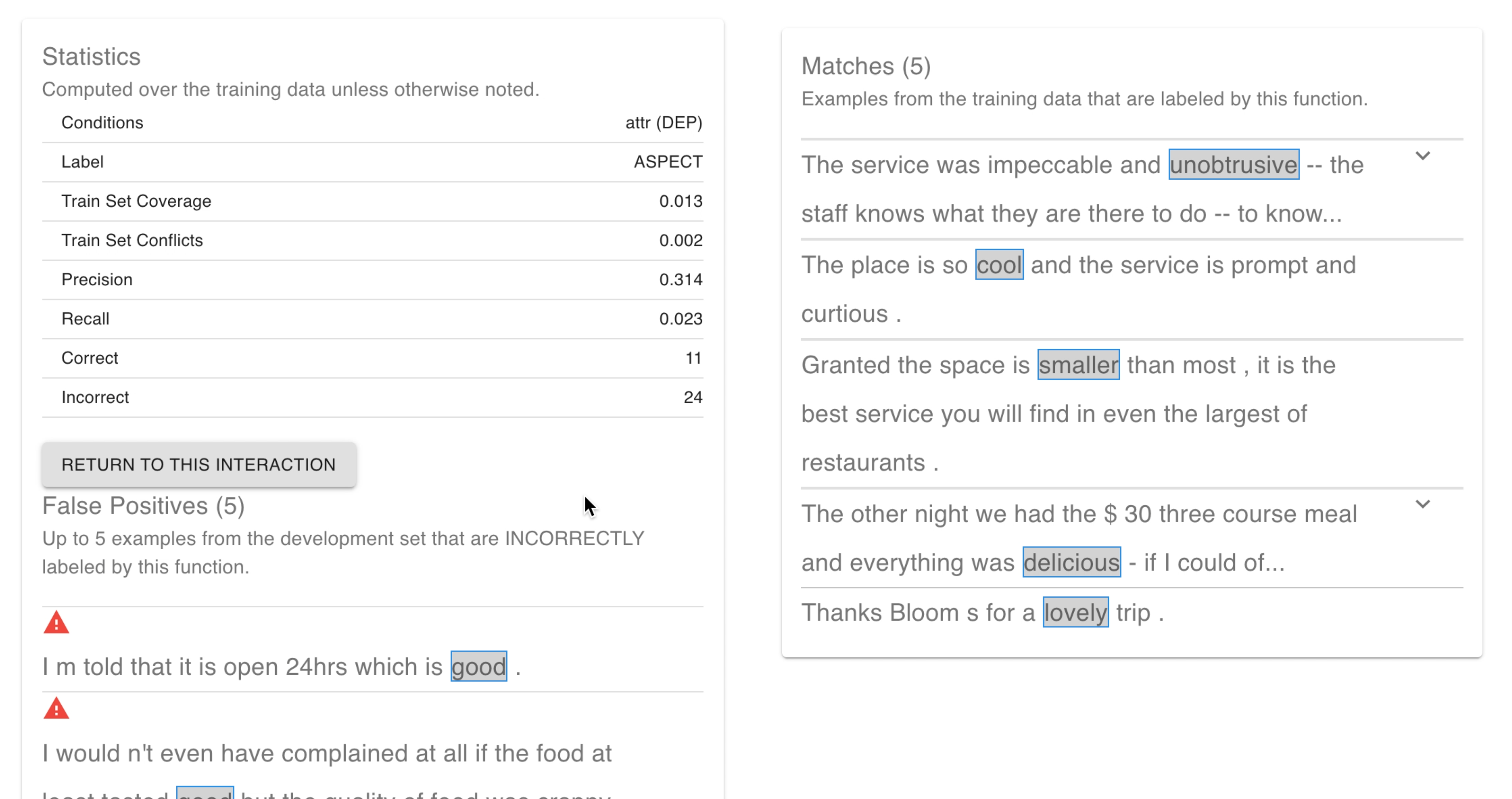}
    \vspace{-2mm}
    \caption{Individual selected labeling function view with false positives and statistics. A user can identify falsely labeled ground truth examples and potentially labeled train examples.}
    \vspace{-2mm}
    \label{fig:false_positives}
\end{figure}
\vspace{-2mm}
\subsection{False Postive Feedback}

TagRuler helps users correct their \textit{mistakes} and develop improved labeling functions.
When a user clicks the $\oi$ button located to the left of each labeling function in the selected labeling function view (\autoref{fig:overview}(C)), false positive feedback and detailed statistics of the corresponding labeling function are displayed as shown in~\autoref{fig:false_positives}.
During our demonstration session, the false positive examples shown by the system were crucial and helped achieving better F1 score and also helped users to identify \textit{wrong ground-truth} instances (meaning that the labeling guideline had changed since the ground truth was labeled, or that the labelers used for the ground truth were not reliable). 


\section{Use Cases and Observations}

In order to demonstrate the efficacy of TagRuler, we demonstrate our findings from <30 minutes trial labeling sessions on two domains.
We recorded the accuracy obtained in these sessions by time and found that TagRuler was superior when compared to a simulated manual labeling session for the same time period.
\subsection{Data and Tasks}
We used two real world datasets: BC5CDR~\cite{wei2015overview}, and Yelp Restaurants~\cite{yelpdataset}. Each dataset was labeled for 30 minutes using TagRuler.
\autoref{tab:dataset-stats} shows more information about the data.
BC5CDR is a medical dataset~\cite{wei2015overview} where each token is annotated as either a chemical, disease, or neither. 
The medical domain is a prime example of a field where training data is expensive or difficult to obtain, as the labeler requires a high level of expertise and their time is very valuable and limited.
This makes it a good use-case for weak-supervision approaches if we can make them accessible.
Yelp Restaurants is a set of restaurant reviews~\cite{yelpdataset} annotated with aspects (ex: food, service) and opinions about those aspects.
Once TagRuler is used to build relevant labeling functions, we can apply the labeling functions to any extension of the datasets and automatically generate training labels.

The labeling tasks of TagRuler were carried out by one of the authors who is not a subject matter expert for these tasks.
We hope that with a little background information about data programming, domain experts would achieve similar results (or better, due to their increased knowledge of the domains).
As a baseline to compare the results of TagRuler's weak labeling and manual labeling, we sampled text examples randomly from the labeled ground truth, at a rate equivalent to the annotator speed for the task. 
We believe that this is a generous estimate, since with TagRuler the user only has to label a few representative tokens, and not every instance in the document.
TagRuler's majority voter model's output and the randomly sampled examples were then used to train CRF models which we evaluate based on their test accuracy scores (micro F1 scores). 

\begin{table}[t]
\sffamily
\centering
\caption{Summary of datasets used for our trial sessions.}
\vspace{-1mm}
\begin{tabular}{lrrp{2.1cm}}
\toprule
 Dataset & \# Doc & Avg. Tokens/Doc & Class Frequency  \\
 \midrule
 BC5CDR &           866 &   381.6 &     Chemical: 0.063 \newline Disease: 0.079 \newline Other: 0.858 \\
 Yelp Restaurants & 838 &   36.5 &      Aspect: 0.100 \newline     Opinion: 0.114 \newline    Other: 0.786 \\
 \bottomrule
\end{tabular}
\vspace{-2mm}
\label{tab:dataset-stats}
\end{table}

\begin{figure}[h]
    \centering
    \vspace{-2mm}
    \includegraphics[width=0.235\textwidth]{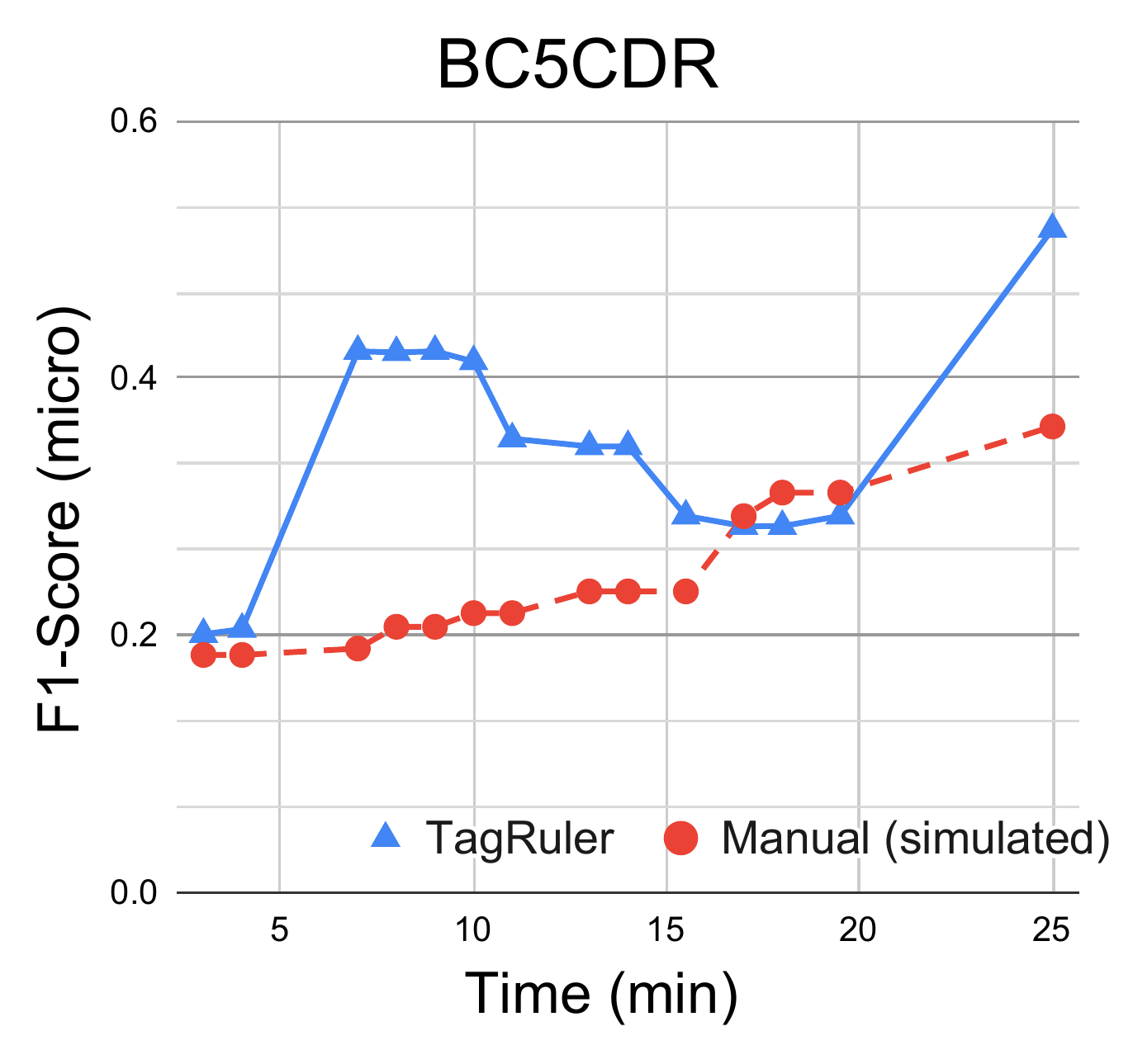}
    \includegraphics[width=0.235\textwidth]{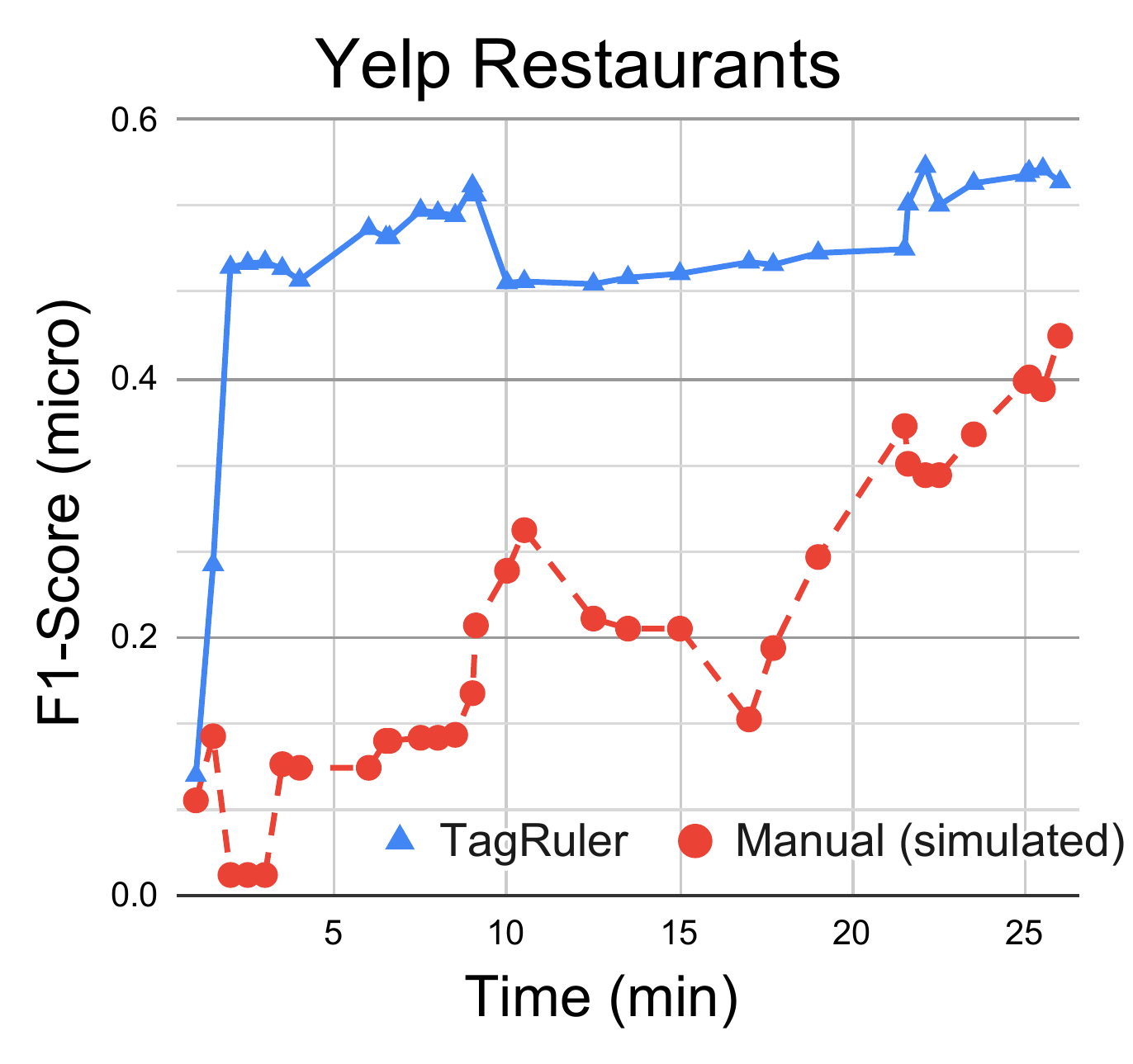}
    \vspace{-4mm}
    \caption{Comparison of TagRuler and simulated manual labeling. Line charts show change in accuracy (micro F1 score) per each text example over time.}
    \vspace{-2mm}
    \label{fig:comparison}
\end{figure}
\vspace{-3mm}
\subsection{Observations}

From less than 30 minutes of labeling BC5CDR dataset, we were able to achieve an F1 score 0.516 from a CRF model trained on TagRuler's generated labels.
With the same time period, the simulated manual labeling ended up an F1 score of 0.363
(\autoref{fig:comparison} left).
On Yelp Restaurants dataset, we achieved an F1 score of 0.564 with a CRF model trained on TagRuler's output in 26 minutes while the simulated manual labeling reached 0.433 in the same time period (\autoref{fig:comparison} right).
The active sampling helped discover some inconsistencies in the ground truth labels. With TagRuler, we don't need to have high confidence in the ground truth labels-- the most reliable ground truth is the domain expert, and TagRuler can leverage this to surface edge cases or changing guidelines.


\section{Related Work}

Both TagRuler and Ruler share concepts and ideas from rule-based information extraction \citep{chiticariu2013rule} and exploit those ideas to make it easier for domain expert to leverage Machine Learning models and data.
This is in contrast to systems like PropMiner \citep{akbik2013propminer}, which focus on generating highly reliable rules for information extraction, as opposed to data programming which relies on a mix of high and low precision rules that make independent errors (like those generated by TagRuler and Ruler). 

The idea of systems and agents that learn by demonstration or by natural language explanation have also been explored in other domains, such as in PUMICE, an agent proposed by \citep{li2019pumice} that uses the combination of natural language programming and programming-by-demonstration to allow users to describe tasks in natural language at a high level, and then starts a collaborative process with the user to recursively resolve ambiguities or vagueness through conversations and demonstrations. Or in the classical artificial intelligence problem of automated synthesis of programs that satisfy a given specification \cite{waldinger1969step}. 

Natural Language Programming has also been applied to Data Programming as in BabbleLabble \citep{hancock2018training}. BabbleLabble allows the labelers to use natural language explanations during the labeling process to generate labeling functions. A semantic parser is then used to convert the natural language explanations into logical forms, which represent labeling functions. The labeling functions are then aggregated using the traditional Data Programming approach proposed in Snorkel \citep{ratner2017snorkel}. We hope TagRuler can complement such an approach by giving users a different, more expressive set of atomic rules.

The difficulty in having Domain Experts capable of generating labeling functions that can go beyond easy and redundant examples is also tackled in SNUBA~\citep{varma2018snuba} and in the framework described in \citep{boecking2020interactive}. In SNUBA, a semi-supervised data programming setting is proposed. But instead of counting on user interactions, SNUBA proposes an automatic generation of labeling functions by using a small set of labeled data, and without user interaction. In \citep{boecking2020interactive}, queries to be annotated are not data points but labeling functions, and the approach is based on training supervised ML models (instead of Data Programming, as done in TagRuler) with weak supervision through an interactive process. 

Finally, span-level annotation has also been explored in the context of Data Programming in the approach proposed in FlyingSquid \citep{fu2020fast}, but as happens in Snorkel, FlyingSquid also requires that the user have programming literacy.
\vspace{-2mm}
\section{Conclusions}

TagRuler is an open-sourced web-based system that enables users to create span-level annotation functions through intuitive interactions.
It requires no programming literacy, yet reliably expands user annotations to expressive rules.
TagRuler's labeling model builds on, and extends, state-of-the-art Data Programming methods, improving user experience and making the techniques more accessible.
Through experiments on two datasets, we demonstrated that an annotator using TagRuler could achieve a higher F1 score compared to manual labeling, allowing faster development of tagging models.
By open-sourcing this technology, we hope to improve the accessibility of deep learning models for NLP.


\bibliographystyle{ACM-Reference-Format}
\bibliography{ref}


\end{document}